# A Delayed Column Generation Strategy for Exact $k$-Bounded MAP Inference in Markov Logic Networks

**Mathias Niepert**
KR & KM Research Group
Universität Mannheim
Mannheim, Germany

## Abstract

The paper introduces $k$-bounded MAP inference, a parameterization of MAP inference in Markov logic networks. $k$-Bounded MAP states are MAP states with *at most $k$* active ground atoms of hidden (non-evidence) predicates. We present a novel delayed column generation algorithm and provide empirical evidence that the algorithm efficiently computes $k$-bounded MAP states for meaningful real-world graph matching problems. The underlying idea is that, instead of solving one large optimization problem, it is often more efficient to tackle several small ones.

## 1 Introduction

Graph matching instances are at the core of many problems in areas such as computational biology, computer vision, and knowledge representation. The problem of aligning protein-protein interaction networks [13] and the problem of matching different heterogeneous ontologies [1] are only two of numerous instances. In several of these cases, the alignment can be derived by computing a MAP state in a probabilistic network whose variables model the potential correspondences [4]. Given an assignment to the observable variables, a (marginal) MAP query attempts to find an assignment to the set of non-evidence (hidden) variables whose score is maximal [3].

We are specifically concerned with computing MAP states in Markov logic networks [5] (MLN) a probabilistic logic which combines the ideas of Markov networks with those of first-order logic. Several scenarios make it necessary to perform exact inference, especially if only those solutions are acceptable that are consistent with respect to the hard formulae of the MLN. Unfortunately, exact MAP inference is NP-hard. In several scenarios, however, it is sufficient to compute a MAP state with at most $k$ active ground atoms. For instance, consider a matching problem where one is interested in the top $k$ consistent correspondences between the nodes of two or more graphs. In other cases, one might know a-priori that only a small fraction of the ground atoms of hidden predicates will be active in a MAP state. This is true for several matching problems where the final alignment is required to be one-to-one and functional. Motivated by these observations, this paper introduces $k$-bounded MAP inference. $k$-Bounded MAP states are MAP states with at most $k$ active ground atoms of hidden predicates. Compare this to the *different $m$-best* MAP problem [2] which aims at finding the $m$ most probable MAP states. We present a novel algorithm combining integer linear programming (ILP) with a form of delayed column generation. Similar to dynamic programming approaches [3] the algorithm inductively builds larger solutions bottom-up from smaller ones. In each iteration, the algorithm solves a restricted version of the original MAP problem (the restricted master problem) and then checks whether the current solution is optimal or if additional variables need to be included in the formulation of the ILP. The latter is the so-called pricing problem or subproblem. We present several technical results concerning the different parts of the algorithm and experimentally evaluate the approach using an established benchmark dataset for ontology matching.

In Section 2 we discuss some concepts such as Markov logic, MAP inference, and the translation to an ILP problem. In Section 3, we define $k$-bounded MAP inference for ML and describe the column generation framework. In Section 4 we introduce the column generation algorithm for exact $k$-bounded MAP inference that we use for our experiments. In Section 5 we present several experimental results supporting our claim that the presented approach efficiently solves MAP inference problems of meaningful real-world applications. In Section 6 we conclude the paper with a brief summary and outlook to future research.

## 2 Foundations

In the following we introduce some foundational concepts such as Markov logic (ML) and integer linear programming (ILP). There is a trade-off between the technical density of the paper and exhaustiveness in defining the formalisms. For well-known concepts such as Markov logic and integer linear programming we omit some technical details and refer the reader to the respective literature on these topics.

### 2.1 Markov Logic

Markov logic combines first-order logic and undirected probabilistic graphical models [5]. A Markov logic network (MLN) is a set of first-order formulae with weights. Intuitively, the more evidence we have that a formula is true the higher the weight of this formula. To simplify the presentation of the technical parts we do *not* include functions. In addition, we assume that all (ground) formulae of a Markov logic network are in clausal form and use the terms *formula* and *clause* interchangeably.

**Syntax**
A signature is a triple $S = (O, H, C)$ with $O$ a finite set of observable predicate symbols, $H$ a finite set of hidden predicate symbols, and $C$ a finite set of constants. A Markov logic network (MLN) is a set of pairs $\{(F_i, w_i)\}$ with each $F_i$ being a function-free first-order formula built using predicates from $O \cup H$ and each $w_i \in \mathbb{R}$ a real-valued weight associated with formula $F_i$. We can represent hard constraints using large weights.

**Semantics**
Let $M = (F_i, w_i)$ be a Markov logic network with signature $S = (O, H, C)$. A *grounding* of a first-order formula $F$ is generated by substituting each occurrence of every variable in $F$ with constants in $C$. Existentially quantified formulae are substituted by the disjunctions of their groundings over the finite set of constants. A formula that does not contain any variables is *ground*. A formula that consists of a single predicate is an *atom*. Note that Markov logic makes several assumptions such as (a) different constants refer to different objects and (b) the only objects in the domain are those representable using the constants [5]. A set of ground atoms is a *possible world*. We say that a possible world $W$ *satisfies* a formula $F$, and write $W \models F$, if $F$ is true in $W$. Let $\mathcal{G}_F^C$ be the set of all possible groundings of formula $F$ with respect to $C$. We say that $W$ satisfies $\mathcal{G}_F^C$, and write $W \models \mathcal{G}_F^C$, if $F$ satisfies every formula in $\mathcal{G}_F^C$. Let $\mathcal{W}$ be the set of all possible worlds with respect to $S$. Then, the probability of a possible world $W$ is given by

$$p(W) = \frac{1}{Z} \exp \left( \sum_{(F_i, w_i)} \sum_{G \in \mathcal{G}_{F_i}^C : \; W \models G} w_i \right).$$

Here, $Z$ is a normalization constant. The score $s_W$ of a possible world $W$ is the sum of the weights of the ground formulae implied by $W$

$$s_W = \sum_{(F_i, w_i)} \sum_{G \in \mathcal{G}_{F_i}^C : \; W \models G} w_i. \qquad (1)$$

### 2.2 MAP Inference and ILP

If we want to determine the most probable state of a MLN, we need to compute the set of ground atoms of the hidden predicates that maximizes the probability given both the ground atoms of observable predicates and all ground formulae. This is an instance of MAP (maximum a-posteriori) inference in the ground Markov logic network. Let $\mathbf{O}$ be the set of all ground atoms of observable predicates and $\mathbf{H}$ be the set of all ground atoms of hidden predicates both with respect to $C$. We make the closed world assumption with respect to the observable predicates. Assume that we are given a set $\mathbf{O}' \subseteq \mathbf{O}$ of ground atoms of observable predicates. In order to find the most probable state of the MLN we have to compute

$$\operatorname*{argmax}_{\mathbf{H}' \subseteq \mathbf{H}} \sum_{(F_i, w_i)} \sum_{G \in \mathcal{G}_{F_i}^C : \; \mathbf{O}' \cup \mathbf{H}' \models G} w_i.$$

In this paper, every $\mathbf{H}' \subseteq \mathbf{H}$ is called a *state*. It is the set of *active* ground atoms of hidden predicates. Markov logic is by definition a declarative language, separating the formulation of a problem instance from the algorithm used for probabilistic inference. MAP inference in Markov logic networks is essentially equivalent to the weighted MAX-SAT problem and, therefore, NP-hard. Integer linear programming (ILP) is an effective method for solving exact MAP inference in undirected graphical models [7, 12] and specifically in Markov logic networks [6]. ILP is concerned with optimizing a linear objective function over a finite number of integer variables, subject to a set of linear constraints over these variables [8]. We omit the formal details of the ILP representation of a MAP problem and refer the reader to [6].

**Example 2.1.** The running example of this paper is a small instance of the ontology alignment problem which involves both soft and hard formulae. ML was successfully applied to ontology matching problems [4]. Let $\mathcal{O}_1$ and $\mathcal{O}_2$ be the two ontologies in Figure 1 with the (a-priori computed) string similarities between the concept labels given in Table 1.

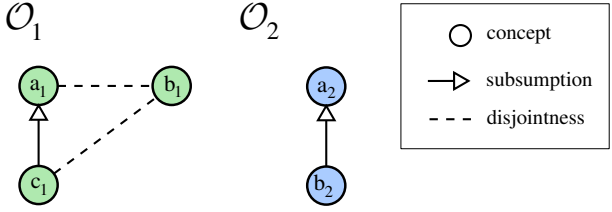

Figure 1: Small fragments of two ontologies.

|    | $a_1$ | $b_1$ | $c_1$ |
|----|-------|-------|-------|
| $a_2$ | 0.95 | 0.25 | 0.12 |
| $b_2$ | 0.55 | 0.91 | 0.64 |

Table 1: A-priori similarities between concept labels.

Let $S = (O, H, C)$ be the signature of a MLN $M$ with $O = \{sub_1, sub_2, dis_1, dis_2\}$, $H = \{map\}$, and $C = \{a_1, b_1, c_1, a_2, b_2\}$. Here, the observable predicates model the subsumption and disjointness relationships between concepts $C$ in the two ontologies and $map$ is the hidden predicate modeling the sought-after matching correspondences. We also assume that the predicates are typed meaning that, for instance, valid groundings of $map(x, y)$ are those with $x \in \{a_1, b_1, c_1\}$ and $y \in \{a_2, b_2\}$. Furthermore, let us assume that the MLN $M$ includes the following formula with weight $w = 10.0$:

$\forall x, x', y, y' :$
$dis_1(x, x') \wedge sub_2(y, y') \Rightarrow (\neg map(x, y) \vee \neg map(x', y'))$

The formula makes those alignments less likely that match concepts $x$ with $y$ and $x'$ with $y'$ if $x$ is disjoint with $x'$ in the first ontology and $y'$ subsumes $y$ in the second. We also include cardinality formulae with weight 10.0 forcing alignments to be one-to-one and functional:

$$\forall x, y, z : map(x, y) \wedge map(x, z) \Rightarrow y = z$$
$$\forall x, y, z : map(x, y) \wedge map(z, y) \Rightarrow x = z$$

In addition, we add the formulae $map(x, y)$ with weight $\sigma(x, y)$ for all $x \in \{a_1, b_1, c_1, d_1\}$ and $y \in \{a_2, b_2\}$ where $\sigma(x, y)$ is the label similarity from Table 1. The observed ground atoms are $sub_1(c_1, a_1), dis_1(a_1, b_1), dis_1(b_1, a_1),$ $dis_1(b_1, c_1), dis_1(c_1, b_1)$ for ontology $\mathcal{O}_1$ and $sub_2(b_2, a_2)$ for ontology $\mathcal{O}_2$. This results in the following relevant ground formulae for the coherence reducing constraint where each observable predicates has been substituted with its observed value:

$$\neg map(a_1, b_2) \vee \neg map(b_1, a_2) \quad (2)$$
$$\neg map(b_1, b_2) \vee \neg map(a_1, a_2) \quad (3)$$
$$\neg map(b_1, b_2) \vee \neg map(c_1, a_2) \quad (4)$$
$$\neg map(c_1, b_2) \vee \neg map(b_1, a_2) \quad (5)$$

For instance, the ground formulae (2) is encoded in an ILP by introducing a new binary variable $y$ which is added to the objective function with coefficient 10.0 and, in addition, by introducing the following linear constraints enforcing the value of $y$ to be equivalent to the truth value of the formula:

$$-x_{a,b} - y \leq -1$$
$$-x_{b,a} - y \leq -1$$
$$x_{a,b} + x_{b,a} + y \leq 2$$

The binary ILP variables $x_{a,b}$ and $x_{b,a}$ correspond to ground atoms $map(a_1, b_2)$ and $map(b_1, a_2)$, respectively. The ILP for our small example includes 19 variables (columns) and 39 linear constraints (12 from the coherence and 27 from the cardinality formulae) which we omit due to space considerations. The preprocessing step of grounding only those clauses that can evaluate to *false* given the current state of observable variables is similar to the approach presented in [9] and was implicitly used in [6].

## 3  $k$-Bounded MAP Inference

In several scenarios it is sufficient to compute a MAP state with Hamming weight at most $k$, that is, with at most $k$ active ground atoms. For instance, consider a matching problem where one is interested in the top $k$ consistent correspondences between nodes of two or more graph structures. Such matching problems occur, for instance, in the areas of computer vision, computational biology, and knowledge representation as we have seen in the previous example. In other cases, one might know a-priori that only a small fraction of the ground atoms of hidden predicates will be active in a MAP state. For instance, this is true for several matching problems where the final matching is required to be one-to-one and functional. In several of these instances it might be necessary to perform exact inference, especially if only those solutions are acceptable that are consistent with respect to the hard formulae of the MLN. In the following we define $k$-bounded MAP states and implement $k$-bounded MAP inference using integer linear programming with a form of delayed column generation.

**Definition 3.1.** Let $M = \{(F_i, w_i)\}$ be a MLN with signature $S = (O, H, C)$. Let $\mathbf{O}' \subseteq \mathbf{O}$ be a set of given ground atoms of observable predicates and let $\hat{\mathbf{H}} \subseteq \mathbf{H}$ be a set of ground atoms of hidden predicates. The $k$-bounded MAP state of $M$ with respect to $\hat{\mathbf{H}}$ is defined as

$$\operatorname*{argmax}_{\substack{\mathbf{H}' \subseteq \hat{\mathbf{H}} \\ |\mathbf{H}'| \leq k}} \sum_{(F_i, w_i)} \sum_{\substack{G \in \mathcal{G}^C_{F_i} : \\ \mathbf{O}' \cup \mathbf{H}' \models G}} w_i. \quad (6)$$

Each $k$-bounded MAP state with respect to $\hat{\mathbf{H}}$ of a ground MLN with $|\mathbf{H}| = N$ is equivalent to a *global* MAP state if $k = N$ and $\hat{\mathbf{H}} = \mathbf{H}$. The underlying idea of our approach is that we compute a *state* with at most $k$ active ground atoms of hidden predicates and verify optimality using the MAP state with at most $k-1$ active ground atoms. The objective is to keep both the number of columns and rows small in the ILP representation. To achieve this, we partition the set of ground atoms of hidden predicates $\mathbf{H}$ in two sets: the *open ground atoms* $\mathbf{H_O}$ and the *closed ground atoms* $\mathbf{H_C}$ of hidden predicates. Intuitively, open ground atoms can potentially be part of the MAP state while closed ground atoms always evaluate to *false*.

### 3.1 Procedural Framework

The proposed $k$-bounded MAP algorithm starts with a small set of $m \geq 1$ open ground atoms of hidden predicates and solves, for each $n \leq k$, the $n$-bounded MAP ILP with respect to the currently *open* ground atoms $\mathbf{H_O}$. This is what is referred to as the *restricted master problem* (RMP) in mathematical programming. Once we have found an $n$-bounded state with respect to $\mathbf{H_O}$ we check for each $h \in \mathbf{H_C}$ whether making $h$ an open ground atom can potentially change the score enough to beat the current solution. This is what is usually called the *subproblem* of column generation algorithms. The optimality of the current $n$-bounded state with respect to all ground atoms of hidden predicates $\mathbf{H}$ is given when none of the ground atoms in $\mathbf{H_C}$ needs to be priced out (moved to $\mathbf{H_O}$) to enter the *basis* of the ILP. To compute the $k$-bounded MAP state with respect to the open ground atoms of hidden predicates $\mathbf{H_O}$ we also construct and solve an ILP. The only additional constraint we have to include is the one restricting the number of active ground atoms of hidden predicates to be less than or equal to $k$. Instead of including all formulae, however, we only need to consider a subset of all ground formulae for the ILP formulation. This is formalized with the following lemma.

**Lemma 3.2.** *Let $M$ be a MLN. Let $\mathbf{H}$ be the set of all ground atoms of hidden predicates, let $\varphi(G_i)$ and $\overline{\varphi}(G_i)$, respectively, be the set of unnegated and negated ground atoms of hidden predicates in ground clause $G_i$, and let $\hat{\mathbf{H}} \subseteq \mathbf{H}$. To compute the $k$-bounded MAP state of $M$ with respect to $\hat{\mathbf{H}}$, the ILP formulation only has to include ground clauses $G_i$ for which the following statement holds*

$$\overline{\varphi}(G_i) \cap (\mathbf{H} \setminus \hat{\mathbf{H}}) = \emptyset.$$

*Proof.* Each ground clause $G_i$ not satisfying the statement holds in every state which is a subset of $\hat{\mathbf{H}}$ and, therefore, need not be part of the ILP formulation. □

The *closed* ground atoms of hidden variables in the set $\mathbf{H_C} = \mathbf{H} \setminus \mathbf{H_O}$ are set to *false* in the ILP formulation. The following example shows how these conditions keep the number of rows and columns small.

**Example 3.3.** Let us revisit Example 2.1. Assume that $\mathbf{H_O} = \{map(a_1, a_2), map(b_1, b_2)\}$ and that we want to compute the 2-bounded MAP state with respect to $\mathbf{H_O}$. By Lemma 3.2 we only have to include the ground formula $\neg map(b_1, b_2) \lor \neg map(a_1, a_2)$ in the formulation of the ILP. The other formulae (1),(3), and (4) from Example 2.1 as well as all cardinality formulae always evaluate to *true*. Instead of 19 variables and 39 rows the ILP here has 3 variables and 3 constraints. The 2-bounded MAP state with respect to $\mathbf{H_O}$ is $\{map(a_1, a_2)\}$.

Given a MLN $M$ and the set of all ground atoms of hidden predicates $\mathbf{H}$, we now have a way to compute, for every $1 \leq k \leq |\mathbf{H}|$, the $k$-bounded MAP state $\mathbf{H}'$ with respect to a set of open ground atoms $\mathbf{H_O} \subseteq \mathbf{H}$. When computing the score of such a $k$-bounded MAP state we only have to add the optimal value of the ILP to the sum of weights of those ground formulae that were excluded by Lemma 3.2. We now need an algorithm to determine whether a $k$-bounded MAP state $\mathbf{H}'$ with respect to $\mathbf{H_O} \subset \mathbf{H}$ is also a $k$-bounded MAP state with respect to $\mathbf{H}$. This optimality test is crucial as it allows us to move from computing the $k$-bounded MAP state to computing the $(k+1)$-bounded MAP state. In the remainder of this section we derive some technical results that will facilitate the desired optimality test.

### 3.2 $k$-Bounded MAP Optimality Test

Most of the following lemmas and definitions are with respect to *ground* MLNs. Again, we assume that all grounded formulae of a given Markov logic network are in clausal form. To simplify the notation, we will write, for a ground formula $G_i$, a set of ground atoms of hidden predicates $\mathbf{H}'$, and a set of given ground atoms of observable predicates $\mathbf{O}'$, $\mathbf{H}' \models G_i$ instead of $\mathbf{O}' \cup \mathbf{H}' \models G_i$ since the set of ground atoms of observable predicates is constant for each MAP instance of a MLN. We first introduce a recursively defined function that maps each *state* to its score.

**Lemma 3.4.** *Let $\mathcal{M} = \{(G_i, w_i)\}$ be a ground MLN, let $\mathbf{H}$ be the set of all ground atoms of hidden predicates in $\mathcal{M}$, let $h \in \mathbf{H}$, and let $\overline{\varphi}(G_i)$ be the set of negated ground atoms of hidden predicates in ground formula $G_i$. The score of a set of ground atoms of hidden predicates is given by the following function $s : 2^{\mathbf{H}} \to \mathbb{R}$*

$$s(\emptyset) = \sum_{\{(G_i, w_i) \mid \overline{\varphi}(G_i) \neq \emptyset\}} w_i \qquad (7)$$

$$s(\mathbf{H}' \cup \{h\}) = s(\mathbf{H}') + c(\mathbf{H}', h) \qquad (8)$$

with $c(\mathbf{H}', h) =$

$$\sum_{\substack{\{(G_i, w_i) \mid \mathbf{H}' \not\models G_i \wedge \\ \mathbf{H}' \cup \{h\} \models G_i\}}} w_i - \sum_{\substack{\{(G_i, w_i) \mid \mathbf{H}' \models G_i \wedge \\ \mathbf{H}' \cup \{h\} \not\models G_i\}}} w_i. \quad (9)$$

**Definition 3.5.** Let $\mathcal{M} = \{(G_i, w_i)\}$ be a ground MLN, let $\mathbf{H}$ be the set of all ground atoms of hidden predicates in $\mathcal{M}$, let $\mathbf{H}' \subseteq \mathbf{H}$ be a set of ground hidden predicates, and let $c : 2^H \times \mathbf{H} \to \mathbb{R}$ be the function defined in Lemma 3.4. We say that a function $\hat{c} : \mathbf{H} \to \mathbb{R}$ $k$-*bounds* $c$ *relative to* $\mathbf{H}'$ if and only if for all $\mathbf{A} \subseteq \mathbf{H}'$ with $|\mathbf{A}| \le k$ and all $h \in \mathbf{H}$ we have that $\hat{c}(h) \ge c(\mathbf{A}, h)$.

Based on this we can state the following theorem.

**Theorem 3.6.** *Let $\mathcal{M} = \{(G_i, w_i)\}$ be a ground MLN, let $\mathbf{H}$ be the set of all ground atoms of hidden predicates, let $s_k$ be the score of a MAP state with at most $k$ active ground atoms of hidden predicates, let $s_{k+1}$ be the score of a state $\mathbf{H}' \subseteq \mathbf{H}$ with at most $k+1$ active ground atoms of hidden predicates, and let $\hat{c}$ be a function that $k$-bounds $c$ relative to $\mathbf{H}$. If for all $h \in \mathbf{H}$ with $\hat{c}(h) > s_{k+1} - s_k$, $h$ was an open ground atom before $\mathbf{H}'$ was computed, then $\mathbf{H}'$ is optimal, that is, $\mathbf{H}'$ is a $(k+1)$-bounded MAP state with respect to $\mathbf{H}$.*

*Proof.* Let us assume that a MAP state with at most $k$ ($k \ge 0$) ground atoms of hidden predicates is given and has score $s_k$ and that $s_{k+1}$ is the score of a state $\mathbf{H}'$ with at most $k+1$ active ground atoms of hidden predicates. Furthermore, let us assume that for all $h \in \mathbf{H}$ with $\hat{c}(h) > s_{k+1} - s_k$, $h$ was an open ground atom before computing $\mathbf{H}'$ but that $\mathbf{H}'$ is *not* optimal. Then, there has to exist a state $\hat{\mathbf{H}}$ with at most $k+1$ ground atoms of hidden predicates that is optimal and an $h' \in \hat{\mathbf{H}}$ with $\hat{c}(h') \le s_{k+1} - s_k$. Since $\mathbf{H}'$ is not optimal but $\hat{\mathbf{H}}$ is optimal, we have $s(\hat{\mathbf{H}}) > s_{k+1}$ and we can infer that $\hat{c}(h') < s(\hat{\mathbf{H}}) - s_k$.
Then, by Lemma 3.4, we can write $\hat{c}(h') < s(\hat{\mathbf{H}}) - s_k$ as $\hat{c}(h') < s(\hat{\mathbf{H}} \setminus \{h'\}) + c(\hat{\mathbf{H}} \setminus \{h'\}, h') - s_k$. Hence, $\hat{c}(h') - c(\hat{\mathbf{H}} \setminus \{h'\}, h') < s(\hat{\mathbf{H}} \setminus \{h'\}) - s_k$. Since $\hat{c}$ $k$-bounds $c$ relative to $\mathbf{H}$ we know that $\hat{c}(h') - c(\hat{\mathbf{H}} \setminus \{h'\}, h') \ge 0$ and, therefore, $s(\hat{\mathbf{H}} \setminus \{h'\}) > s_k$. Hence, $\hat{\mathbf{H}} \setminus \{h'\}$ is a MAP state with at most $k$ ground atoms of hidden predicates and $s(\hat{\mathbf{H}} \setminus \{h'\}) > s_k$, a contradiction to our assumption that $s_k$ is an optimal score. □

Theorem 3.6 is the basis of the proposed method for verifying the optimality of a state with at most $k+1$ active ground atoms using the optimality of a state with at most $k$ active ground atoms both relative to $\mathbf{H}$. An effective optimality test requires an algorithm computing a function that $k$-bounds $c$ relative to $\mathbf{H}$ for all potential $k$ as tightly as possible. The computation of this function should be efficient and avoid the inclusion of many ground formulae. In the next definition we construct a *subproblem* whose optimal value $k$-bounds $c$ relative to $\mathbf{H}$.

**Definition 3.7.** Let $\mathcal{M} = \{(G_i, w_i)\}$ be a ground MLN, let $\mathbf{H}$ be the set of all ground atoms of hidden predicates, let $\mathbf{H}' \subseteq \mathbf{H}$ be the set of *open* ground atoms, let $\varphi(G_i)$ and $\overline{\varphi}(G_i)$ be the set of unnegated and negated ground atoms in ground formula $G_i$, let $k \in \mathbb{N}$, and let $h \in \mathbf{H}'$. The $k$-*subproblem of $h$ relative to $\mathbf{H}'$*, which is also an ILP, is constructed as follows:

1. For each $G_i$ with (a) $h \in \varphi(G_i)$; (b) $h \notin \overline{\varphi}(G_i)$; and (c) $\overline{\varphi}(G_i) \cap (\mathbf{H} \setminus \mathbf{H}') = \emptyset$ we introduce a binary variable $x_j$ with coefficient $c_j = w_i$ and add the linear constraints equivalent to $\neg \tilde{G}_i \Leftrightarrow x_j$, where $\tilde{G}_i$ is the clause $G_i$ where all occurrences of $h$ have been removed;

2. For each $G_i$ with (a) $h \in \overline{\varphi}(G_i)$; (b) $h \notin \varphi(G_i)$; and (c) $\overline{\varphi}(G_i) \cap (\mathbf{H} \setminus \mathbf{H}') = \emptyset$ we introduce a binary variable $x_j$ with coefficient $c_j = -w_i$ and add the linear constraints equivalent to $\neg \tilde{G}_i \Leftrightarrow x_j$, where $\tilde{G}_i$ is the clause $G_i$ where all occurrences of $h$ have been removed;

3. We add one constraint restricting the number of active ground atoms of hidden predicates to be less than or equal to $k$.

The objective of the ILP is then **Maximize** $\sum_j c_j x_j$.

The formulation of the subproblem can be adjusted to achieve certain computational goals. For instance, if we wanted to achieve a speed-up of the subproblem computation we could remove the constraint that all variables are integer. The optimal value of the resulting subproblem would then be an upper-bound to the optimal value of the original formulation.

**Example 3.8.** We revisit Example 2.1. Assume that we want to compute the 1-*subproblem* of $map(a_1, b_2)$ relative to $\mathbf{H}$. $map(a_1, b_2)$ occurs unnegated only in the ground formula $map(a_1, b_2)$ with weight 0.55. Hence, by rule (1), we add a new binary variable $x_1$ to the ILP with coefficient 0.55 and no constraint because $true \Leftrightarrow x_1$ is equivalent to $x_1$. Ground atom $map(a_1, b_2)$ occurs negated in the ground formula $\neg map(a_1, b_2) \vee \neg map(b_1, a_2)$ with weight 10.0. Hence, by rule (2), we add a new binary variable $x_2$ with coefficient $-10.0$ to the ILP and the constraints $x_2 - x_{b,a} \le 0$ and $x_{b,a} - x_2 \le 0$ which are equivalent to $map(b_1, a_2) \Leftrightarrow x_2$. The same is done for three more cardinality formulae involving $map(a_1, b_2)$, introducing variables $x_3, x_4$, and $x_5$. Finally, we add the constraint $x_{a,a} + x_{b,a} + x_{b,b} + x_{c,a} \le 1$ by rule (3). The optimal value of this ILP is 0.55.

The following theorem states the desired optimality of the *subproblem*.

**Theorem 3.9.** *Let $\mathcal{M} = \{(G_i, w_i)\}$ be a ground MLN, let $\mathbf{H}$ be the set of all ground atoms of hidden predicates, let $\mathbf{H}' \subseteq \mathbf{H}$ be the set of open ground atoms, let $\varphi(G_i)$ and $\overline{\varphi}(G_i)$ be the set of unnegated and negated ground atoms in ground formula $G_i$, let $k \in \mathbb{N}$, let $h \in \mathbf{H}$, let $c$ be the function from Lemma 3.4, and let $o_h$ be the optimal value of the k-subproblem of $h$ relative to $\mathbf{H}'$. Then,*

$$o_h = \max_{\{\mathbf{A} \subseteq \mathbf{H}',\ |\mathbf{A}| \leq k\}} c(\mathbf{A}, h).$$

*Proof sketch.* We provide a sketch of the proof by showing that every $\mathbf{A}' \subseteq \mathbf{H}'$ with $|\mathbf{A}'| \leq k$ corresponds to a feasible solution of the k-subproblem of $h$ relative to $\mathbf{H}'$ with value $c(\mathbf{A}', h)$. Since the ILP variables representing the elements in $\mathbf{A}'$ do not overlap with the variables $x_j$ and since $|\mathbf{A}'| \leq k$ we know that every linear constraint is satisfied. For each $G_i$ with $h \in \varphi(G_i)$, we have that either (a) the corresponding ILP variable $x_j$ is part of the feasible solution or (b) $x_j$ is *not* part of the feasible solution. In case (a) we have $\mathbf{A}' \not\models G_i$ and $\mathbf{A}' \cup \{h\} \models G_i$, and in case (b) we have $\mathbf{A}' \models G_i$ and $\mathbf{A}' \cup \{h\} \models G_i$. Furthermore, for each $G_i$ with $h \in \overline{\varphi}(G_i)$, we have either (a) the corresponding ILP variable $x_j$ is part of the feasible solution or (b) $x_j$ is *not* part of the feasible solution. In case (a) we have $\mathbf{A}' \models G_i$ and $\mathbf{A}' \cup \{h\} \not\models G_i$, and in case (b) we have $\mathbf{A}' \models G_i$ and $\mathbf{A}' \cup \{h\} \models G_i$. Furthermore, for any set $\mathbf{A}' \subseteq \mathbf{H}'$ and each ground clause $G_i$ we have $\mathbf{A}' \not\models G_i$ and $\mathbf{A}' \cup \{h\} \models G_i$ only if (a) $h \in \varphi(G_i)$; (b) $h \notin \overline{\varphi}(G_i)$; and (c) $\overline{\varphi}(G_i) \cap (\mathbf{H} \setminus \mathbf{H}') = \emptyset$; and for each ground clause $G_i$ we have $\mathbf{A}' \models G_i$ and $\mathbf{A}' \cup \{h\} \not\models G_i$ only if (a) $h \in \overline{\varphi}(G_i)$; (b) $h \notin \varphi(G_i)$; and (c) $\overline{\varphi}(G_i) \cap (\mathbf{H} \setminus \mathbf{H}') = \emptyset$. Hence, by Lemma 3.4, we conclude that $c(\mathbf{A}', h)$ is the value of the feasible ILP solution corresponding to $\mathbf{A}'$. □

## 4 Algorithm

Algorithm 1 brings the different components together. First, we need to find a strategy for choosing the $m$ ground atoms of hidden predicates to be priced out (moved to $\mathbf{H_O}$ in lines 7 and 18). Several alternatives are possible. One is choosing the ground atoms with the $m$ highest optimal values $o_h$ of the respective subproblems. In our experiments we used the a-priori weights of the individual ground atoms. For each $n \leq k$ we compute the score $s_n$ of the (n-1)-bounded MAP state $\mathbf{H}'$ with respect to the open ground atoms $\mathbf{H_O}$ (line 12). We then test, for each closed ground atom $h \in \mathbf{H_T}$, whether it could potentially be part of an n-bounded state with respect to $\mathbf{H}$ with higher

**Algorithm 1** Compute k-bounded MAP state

**Require:** $1 \leq k, m \leq |\mathbf{H}|$
1: **Input:** MLN $M$ with signature $S = (O, H, C)$ and parameter $k$ and $m$
2: $\mathbf{H_O} \leftarrow \emptyset$
3: $\mathbf{H_C} \leftarrow \mathbf{H}$
4: compute score $s_0$ of MAP state $\mathbf{H}'$ with respect to $\mathbf{H_O}$ subject to $|\mathbf{H}'| \leq 0$ using Lemma 3.4 (7)
5: $n \leftarrow 1$
6: $s_{n-1} \leftarrow s_0$
7: move $m$ ground atoms from $\mathbf{H_C}$ to $\mathbf{H_O}$
8: **repeat**
9:    compute score $s_n$ of MAP state $\mathbf{H}'$ with respect to $\mathbf{H_O}$ subject to $|\mathbf{H}'| \leq n$
10:    $\mathbf{H_T} \leftarrow \mathbf{H_C}$
11:    **for all** $h \in \mathbf{H_T}$ **do**
12:      compute the optimal value $o_h$ of the (n-1)-subproblem of $h$ relative to $\mathbf{H_T} \cup \mathbf{H_O}$
13:      **if** $o_h \leq s_n - s_{n-1}$ **then**
14:        $\mathbf{H_T} \leftarrow \mathbf{H_T} \setminus \{h\}$
15:      **end if**
16:    **end for**
17:    **if** $\mathbf{H_T} \neq \emptyset$ **then**
18:      move $m$ ground atoms from $\mathbf{H_C}$ to $\mathbf{H_O}$
19:    **else**
20:      $s_{n-1} \leftarrow s_n$
21:      $n \leftarrow n + 1$
22:    **end if**
23: **until** $n = k + 1$
24: **return** $\mathbf{H}'$

score (line 13). If this is the case, we move $m$ closed ground atoms to the set of open ground atoms (line 18) and repeat the previous steps. Otherwise, the optimality of $\mathbf{H}'$ with respect to $\mathbf{H}$ is guaranteed by Theorem 3.6 and Theorem 3.9, and we continue with computing the n+1-bounded MAP state. When constructing the ILPs for (a) the restricted master problem (line 9) and (b) the subproblems (line 12) we have to retrieve ground formulae from the MLN representation each time, which is potentially very time consuming. As proposed similarly in [6], however, this problem can be transformed into a database query evaluation problem. For instance, when we want to retrieve all ground formulae $G_i$ for which $h \in \varphi(G_i)$ we can translate this into a conjunctive query over the appropriately indexed relational database.

**Example 4.1.** Let us revisit Example 2.1 once more. We first compute the score $s_0$ of the 0-bounded MAP state (line 4). This, of course, is the sum of the weights of ground clauses with at least one negated ground atom (see Equation (7) in Lemma 3.4). Example 2.1 has 13 such ground formulae with weight 10.0. Hence, $s_0 = 130.0$. Here, we choose

$m = 1$ and move $map(a_1, a_2)$ to $\mathbf{H_O}$ since 0.95 is the highest a-priori weight. We then compute the score $s_1$ of the 1-bounded MAP state with respect to $\mathbf{H_O} = \{map(a_1, a_2)\}$ which is 130.95. For each closed ground atom $h$ we compute the optimal value $o_h$ of the 0-subproblem of $h$ relative to $\mathbf{H_T} \cup \mathbf{H_O}$. For each of these $o_h \leq s_1 - s_0 = 0.95$ and, therefore, $\{map(a_1, a_2)\}$ is a 1-bounded MAP state. We move $map(b_1, b_2)$ to $\mathbf{H_O}$ since 0.91 is the second highest a-priori weight. When computing the 2-bounded MAP state with respect to $\mathbf{H_O} = \{map(a_1, a_2), map(b_1, b_2)\}$ the coherence formula (3) causes the result to be $\{map(a_1, a_2)\}$ with score 130.95. Since $s_2 - s_1 = 0$ the optimality of the current state with respect to $\mathbf{H}$ cannot be verified. Hence, $map(c_1, b_2)$ is moved to $\mathbf{H_O}$ since 0.64 is the third highest weight. The score of the 2-bounded MAP state with respect to $\mathbf{H_O}$ is $130.0 + 0.95 + 0.64$ and, therefore, $s_2 - s_1 = 0.64$. For each closed ground atom we compute $o_h$ for the 1-subproblem relative to $\mathbf{H_T} \cup \mathbf{H_O}$. For each of these it is $o_h \leq s_2 - s_1 = 0.64$ and, therefore, $\{map(a_1, a_2), map(c_1, b_2)\}$ is a 2-bounded MAP state. In addition, $\{map(a_1, a_2), map(c_1, b_2)\}$ is the global MAP state for the MLN since a one-to-one and functional alignment of $\mathcal{O}_1$ and $\mathcal{O}_2$ contains at most two ground atoms. Instead of translating the problem into one ILP instance we computed the MAP state by solving several smaller ILP instances.

## 5 Experiments

We used the Ontofarm dataset [11] for our experiments. It is the dataset of the OAEI[1] conference track and consists of several ontologies modeling the domain of academic conferences. The ontologies were designed by different groups and, therefore, reflect different conceptualizations of the same domain. Reference alignments (gold standard alignments) for seven of these ontologies are made available by the organizers. These 21 alignments contain correspondences between concepts and properties including a large number of non-trivial cases. From the 7 ontologies with reference alignment we chose the EKAW (77 concepts), SIGKDD (49 concepts), ConfTool (38 concepts), and CMT (36 concepts) ontologies. The signature and formulae of the MLN are identical with those introduced in Example 2.1 except that we added additional formulae with weight 0.1 making alignments that match $x$ with $y$ and $x'$ with $y'$ more likely if $x'$ subsumes $x$ in $\mathcal{O}_1$ and $y'$ subsumes $y$ in $\mathcal{O}_2$:

$\forall x, x', y, y'$:
$sub_1(x, x') \land sub_2(y, y') \Rightarrow (map(x, y) \Rightarrow map(x', y'))$.

For the label similarity $\sigma$ we decided to use a standard lexical similarity measure. Thus, after converting the concept names to lowercase and removing delimiters and stop-words, we applied a string similarity measure based on the Levensthein distance. We applied the reasoner Pellet [10] to create the ground MLN formulation and the mixed integer programming solver SCIP[2] to solve the ILPs. We then applied our algorithm to compute the $k$-bounded MAP states for $1 \leq k \leq 20$ and with parameter $m = 10$ using the label similarity as pricing criterion. The experiments were conducted on a PC with an AMD Athlon dual core 5400B 1.0 GHz processor and 3 GB RAM.

[1] http://oaei.ontologymatching.org

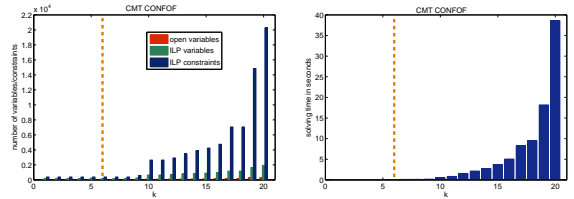

(a) cmt-confof: ILP dimensions
(b) cmt-confof: solving time in seconds

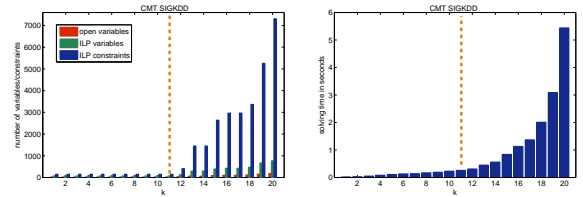

(c) cmt-sigkdd: ILP dimensions
(d) cmt-sigkdd: solving time in seconds

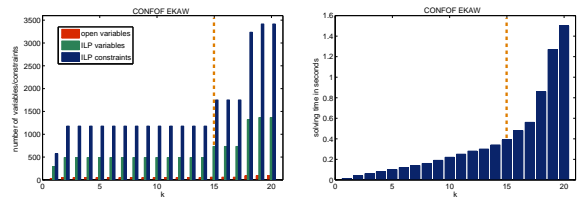

(e) confof-ekaw: ILP dimensions
(f) confof-ekaw: solving time in seconds

Figure 2: ILP dimensions and *accumulative* running times of the ILP solver for several ontology matching instances. The dotted (orange) lines indicate the value of $k$ at which the F1-value attains its maximum.

Figure 2 depicts the results for three different matching instances. The ILP dimensions, that is, the number of columns and rows and the accumulative solving times remain small for $k \leq 15$. The running time spent on the subproblem ILPs was always less than 10% of the overall computation time. Remarkably, the $k$-bounded MAP states leading to the respective alignments with the highest $F_1$ scores (the harmonic mean of precision and recall determined using the gold standard) can be computed very efficiently. Table 2 compares the solving times and the ILP dimensions of the column generation algorithm with the naïve approach which constructs the entire ILP with the additional cardinality constraint. The column generation approach is more than 5000 times faster than the naïve approach to compute the top 10 alignment between the ConfOf and EKAW ontologies. The computation times of less than one second, for instance, would allow real-time user interaction with an alignment system.

[2] http://scip.zib.de/

| **time** | cmt-confof | cmt-sigkdd | confof-ekaw |
|---|---|---|---|
| naïve | 167.08 | 144.0 | 1181.38 |
| CG | 0.51 | 0.23 | 0.22 |
| **dims** | cmt-confof | cmt-sigkdd | confof-ekaw |
| naïve | 5566/194340 | 4085/318151 | 35758/869636 |
| CG | 645/2643 | 72/149 | 489/1178 |

Table 2: Accumulative solving times [in seconds] and ILP dimensions [columns/rows] of the naïve and the column generation (CG) algorithm to compute the 10-bounded MAP state for three different ontology alignment instances.

## 6  Conclusion

We presented $k$-bounded MAP inference and motivated it by use-cases from the areas of computational biology and knowledge representation. It is a useful concept in the context of graph matching problems where one is interested in the top $k$ correspondences between nodes of two or more graph structures. The presented column generation algorithm is tailored to the problem and is especially efficient for instances where (a) a-priori weights are given for ground atoms or (b) it is known a-priori that the number of active ground atoms in a MAP state is small relative to the number of all ground atoms of hidden predicates. Typical instances are one-to-one and functional alignment problems. The approach also lends itself to distributed computing strategies since the individual subproblems are mutually independent. We will try to leverage existing software platforms for distributed computing to implement distributed probabilistic reasoning. The presented algorithm can also be transformed into an approximate inference algorithm by adjusting the optimality test. Future research might be concerned with applying the presented ideas to different graph matching as well as weighted MAX-SAT problems.

**Acknowledgments**

Many thanks to Christian Meilicke, Sebastian Riedel, and Heiner Stuckenschmidt for helpful discussions and the anonymous reviewers for valuable feedback.